%% file: aaai25.tex
\documentclass[letterpaper]{article} 
\usepackage{aaai25}  
\usepackage{times}  
\usepackage{helvet}  
\usepackage{courier}  
\usepackage[hyphens]{url}  
\usepackage{graphicx} 
\usepackage{natbib}  
\usepackage{caption} 
\usepackage{xcolor}
\definecolor{darkgreen}{rgb}{0,0.5,0}
\usepackage{soul}
\usepackage{pifont}
\usepackage{xcolor}
\usepackage{multirow}
\usepackage{algorithm}
\usepackage{algorithmic}
\usepackage{booktabs}
\usepackage{amssymb}
\usepackage{subcaption}
\usepackage{newfloat}
\usepackage{listings}
\DeclareCaptionStyle{ruled}{labelfont=normalfont,labelsep=colon,strut=off} 
\floatstyle{ruled}
\newfloat{listing}{tb}{lst}{}
\floatname{listing}{Listing}
\title{Tool-Assisted Agent on SQL Inspection and Refinement in Real-World Scenarios}
\author{
    Zhongyuan Wang\textsuperscript{\rm 1},
    Richong Zhang\textsuperscript{\rm 1,2 \thanks{Corresponding author}},
    Zhijie Nie\textsuperscript{\rm 1,3},
    Jaein Kim\textsuperscript{\rm 1}
}
\affiliations{
    \textsuperscript{\rm 1}CCSE, School of Computer Science and Engineering, Beihang University, Beijing, China\\
    \textsuperscript{\rm 2}Zhongguancun Laboratory, Beijing, China\\
    \textsuperscript{\rm 3}Shen Yuan Honors College, Beihang University, Beijing, China\\
    \{wangzy23,zhangrc,niezj\}@act.buaa.edu.cn, jaein@buaa.edu.cn
}

\usepackage{bibentry}

\newcommand{\ourdataset}{Spider-Mismatch}
\newcommand{\ourmethod}{Tool-SQL}

\begin{document}
\nocopyright
\maketitle

\input{0-Abstract}
\input{1-Introduction}
\input{2-RelatedWorks}
\input{3-Method}

\input{3.5-Dataset}

\input{4-Experiments}
\input{5-Conclusion}

\bibliography{aaai25}

\end{document}

%% file: 0-Abstract.tex
\begin{abstract}
Recent Text-to-SQL methods leverage large language models (LLMs) by incorporating feedback from the database management system. While these methods effectively address execution errors in SQL queries, they struggle with database mismatches---errors that do not trigger execution exceptions. Database mismatches include issues such as condition mismatches and stricter constraint mismatches, both of which are more prevalent in real-world scenarios. To address these challenges, we propose a tool-assisted agent framework for SQL inspection and refinement, equipping the LLM-based agent with two specialized tools: a retriever and a detector, designed to diagnose and correct SQL queries with database mismatches. These tools enhance the capability of LLMs to handle real-world queries more effectively. We also introduce \ourdataset, a new dataset specifically constructed to reflect the condition mismatch problems encountered in real-world scenarios. Experimental results demonstrate that our method achieves the highest performance on the averaged results of the Spider and Spider-Realistic datasets in few-shot settings, and it significantly outperforms baseline methods on the more realistic dataset, \ourdataset.
\end{abstract}

%% file: 1-Introduction.tex
\section{Introduction}

The Text-to-SQL task \cite{zhong2017seq2sql, yu2018spider} aims to automatically convert natural language questions from users into Structured Query Language (SQL) queries with the database schema, enabling non-expert users to more easily access data from databases.

Previous works on the Text-to-SQL task \cite{wang2020rat, scholak-etal-2021-picard, gan2021towards} focus on training models using various frameworks and strategies, which typically require a large number of labeled data. Recent research explores the power of large language models (LLMs) and applies the in-context learning (ICL) paradigm to this task. In-context learning is an emerging capability of LLMs that allows them to perform comparably to fine-tuned models on many tasks simply by giving them a few examples. Initial approaches to in-context learning \cite{zhang2023act, dong2023c3} focus on creating better prompts to leverage the single-step reasoning capabilities of LLMs. Later approaches \cite{pourreza2024din, wang2024macsql} introduce multi-step processes to assist LLMs in SQL query generation. Given the challenges in producing entirely correct SQL queries, existing methods incorporate a refinement stage, which can be divided into two types: self-correction and refinement based on execution feedback. The self-correction approach \cite{pourreza2024din} guides LLMs to generate revised SQL queries based on predefined correction guidelines, though it can only address a limited range of errors. In contrast, the execution feedback approach \cite{wang2024macsql} refines SQL queries by leveraging feedback from executing these queries on a database management system (DBMS), ensuring excitability and improving the results. 

\begin{figure}[t]
    \centering
    \includegraphics[width=0.98\columnwidth]{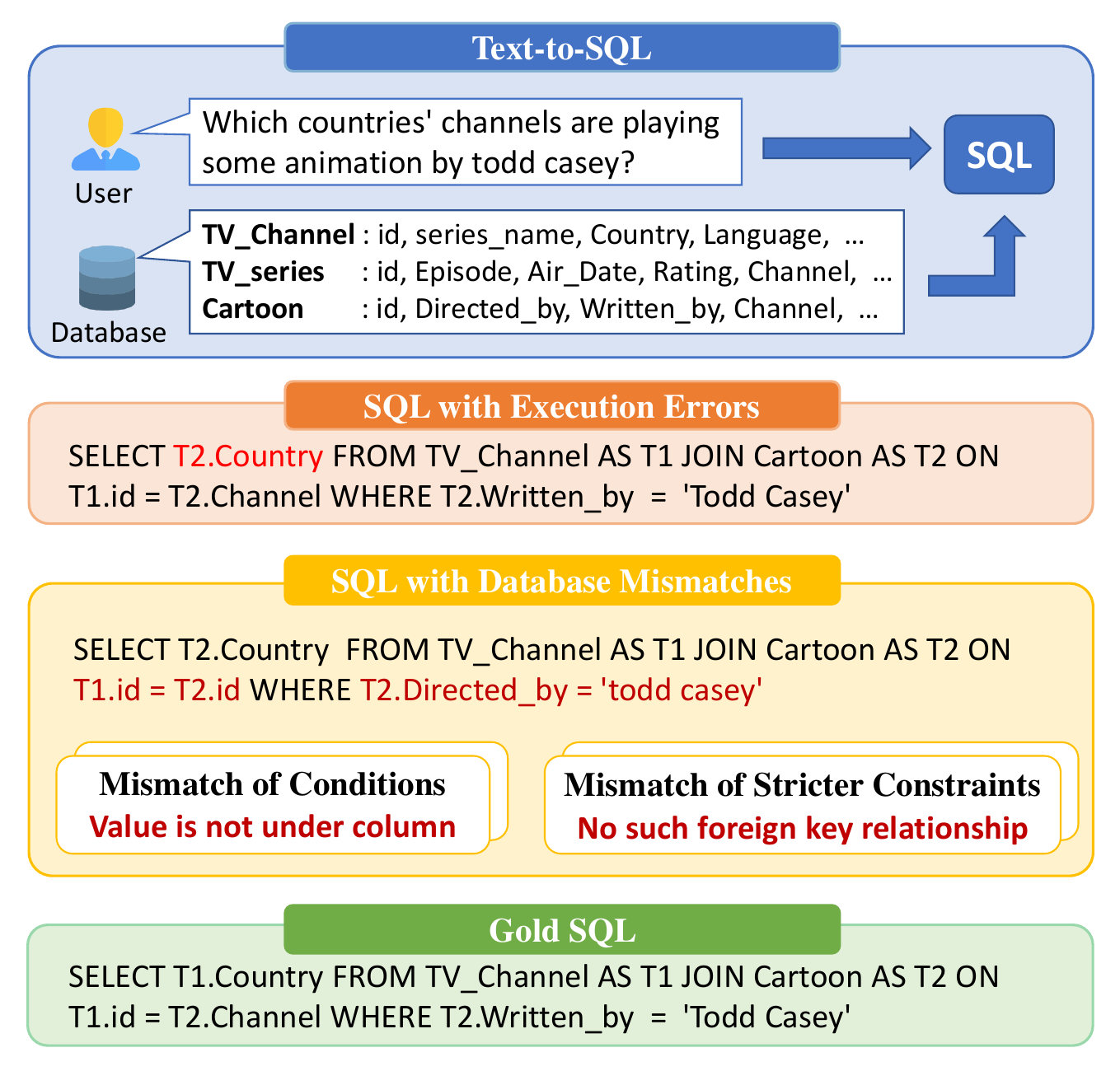}
    \caption{A real-world example of the Text-to-SQL task. Compared to \textbf{execution errors}, diagnosing \textbf{database mismatches} in SQL queries is more challenging.}
    \label{fig:introduction}
\end{figure}

Although existing methods address execution errors in SQL queries by leveraging feedback from the DBMS, they struggle with issues that do not trigger execution exceptions. We focus on a specific type of error within this category, which we term \textbf{database mismatches}. This error type encompasses two common and challenging problems in real-world scenarios. (1) \textbf{Mismatch of Conditions:} 
The mismatch of conditional clauses in SQL queries can lead to either empty or incorrect results. 
In real-world scenarios, the diverse and irregular nature of user questions makes it difficult for LLMs to accurately align questions with the database and generate correct SQL condition clauses. 
Figure \ref{fig:introduction} illustrates the challenges of translating a natural language query into SQL in real-world scenarios, emphasizing the ambiguity often present in user questions. Such ambiguity complicates the LLMs to accurately determine whether the user is referring to specific columns like ``Written\_by" or ``Directed\_by" due to the lack of clarity in the question. Furthermore, even if the correct columns are identified, inconsistencies between the values mentioned in the user question and the actual data in the database can lead to empty results (e.g., ``todd casey" vs. ``Todd Casey"). Although existing methods attempt to assist LLMs by providing example values for each column, which is insufficient in real-world scenarios.
(2) \textbf{Mismatch of Stricter Constraints:} 
In real-world scenarios, SQL queries often need to adhere to stricter constraints, which may stem from the inherent features of SQL or user-defined rules. For instance, the former might involve restrictions related to foreign key relationships or column data types, and the latter could include mandatory processes such as  ``NULL" values or specific data formats. These mismatches of stricter constraints are not reflected during execution, but SQL queries that do not satisfy these constraints may not yield the expected results. 

These challenges make it difficult for LLMs to generate accurate SQL queries in a single process, requiring a multi-step refinement to accomplish SQL generation. 
To enhance the SQL generation capabilities of LLMs in real-world scenarios, we propose \ourmethod, a tool-assisted agent framework designed to continuously inspect and correct errors in SQL queries. Our framework employs various tools to diagnose problems within SQL queries and leverages an LLM-based agent to purposefully refine these queries based on the feedback provided by these tools. 
We design two specific tools to address the aforementioned problems: (1) \textit{Database Retriever}, which helps LLM-based agents by retrieving similar database cells as feedback when SQL conditional clauses do not match any entries in the databases. (2) \textit{Error Detector}, which diagnoses a wider range of errors, including execution errors and mismatches of stricter constraints defined by SQL rules or domain experts. 

Additionally, we observe that the mainstream Spider dataset and its variants \cite{yu2018spider, gan2021towards, deng2021structure, gan2021exploring} rarely reflect the mismatches of conditions in SQL queries, which usually requires verification in databases. 
Meanwhile, the values mentioned in the conditions of most questions in existing datasets (including the Bird dataset \cite{li2024can}) are identical to corresponding cells in databases, which is rare in the real world.
To bridge the gap between existing datasets and real-world scenarios, we introduce \ourdataset, a new dataset specifically designed to highlight the mismatch problem in SQL conditional clauses. We modify the questions and corresponding SQL queries in Spider and its variants by applying specific disturbances, which challenge models to generate accurate SQL queries.

Our main contributions can be summarized as follows:
\begin{itemize}
	\item We propose \ourmethod, a tool-assisted agent framework aiming to inspect problems in SQL queries with the assistance of specialized tools and refine these queries using an LLM-based agent. 
	\item We introduce \ourdataset, a new dataset designed specifically to bridge the gap between existing benchmarks and real-world scenarios, where the semantics of the questions are more ambiguous.
	\item \ourmethod\ achieves the highest execution accuracy on the overall results of Spider and Spider Realistic in few-shot settings. Experiments on \ourdataset\ demonstrate that our method maintains strong performance despite the increased ambiguity in question narratives encountered in real-world scenarios.
\end{itemize}

%% file: 2-RelatedWorks.tex
\section{Related Work}
\subsection{LLM for Text-to-SQL}
LLMs have been widely used in the Text-to-SQL task, with various approaches proposed to enhance their ability. 
Some research focuses on designing higher-quality prompts for LLMs to explore the potential of LLMs in Text-to-SQL parsing. For example, ACT-SQL \cite{zhang2023act} and \cite{tai-etal-2023-exploring} enhance LLM's reasoning capabilities through chain-of-thought prompts. The DAIL-SQL \cite{gao2024text} systematically investigates prompt engineering in LLM Text-to-SQL, including question representation, demonstration selection, and demonstration organization. 
Most recent research employs multi-stage frameworks, aiming to enhance the performance of LLMs by decomposing the Text-to-SQL task into smaller sub-tasks and designing different prompts for each sub-task.  
For instance, DIN-SQL \cite{pourreza2024din} decomposes the Text-to-SQL task into schema linking, question classification, SQL generation, and self-correlation to reduce the overall difficulty. DEA-SQL \cite{xie2024decomposition} enhances the process of DIN-SQL and introduces an additional active learning module.
To reduce errors in generated SQL queries, existing multi-stage methods often introduce error correction modules. DIN-SQL and DEA-SQL adopt self-correction, guiding LLMs to correct the SQL based on the static guidelines in prompts. MAC-SQL \cite{wang2024macsql} leverage the feedback from the database management system to guide the LLM, solving the execution errors in SQL queries.

\subsection{LLM-based Agents}
With the rise of LLM, the potential of LLM-based agents is continuously being explored. Invoking tools is a crucial capability for LLM agents, which bridge the gap between LLM agents and the external world. AutoGPT \cite{autogpt2023} is an open-source implementation of the AI agent, with many useful tools to augment a single agent. OpenAgents \cite{xie2023openagents} develops three agents, each specializing in different domains and equipped with domain-specific tools.
ToolLLM \cite{qin2024toolllm} and API-Bank \cite{li2023api} focus on LLM Agent interacting with a wide range of open-domain real-world applications with RESTful APIs.

In the Text-to-SQL task, MAC-SQL \cite{wang2024macsql} proposed a multi-agent framework that separately addresses various sub-tasks of Text-to-SQL, including SQL refinement through execution exceptions. However, leveraging tools to diagnose the other types of errors in SQL queries and provide feedback to assist LLM-based agents in performing SQL refinement is under-explored. We fill this gap and explore the use of tools to inspect and address the database mismatches in SQL queries.

%% file: 3-Method.tex
\section{Method}

\subsection{LLM-based Text-to-SQL Task}
LLM-based methods typically adopt the in-context learning paradigm, treating Text-to-SQL as a generation task. The generation process can be formulated as:
\begin{equation}
    Y = f_{\rm LLM}(I,E,S,Q),
\end{equation}
where the input to the large language models $f_{\rm LLM}$ includes a task instruction prompt $I$, a set of demonstration examples $E$, a database schema $S$ of the database $D$, and a new query $Q$. The demonstrations $E=[(S_1, Q_1, Y_1), ..., (S_k, Q_k, Y_k)]$ consists of $k$ examples from the training set, each with expected output $Y_i$. The output $Y$ from the LLM can be either an SQL query or intermediate results in other forms.

\subsection{Framework}

\begin{figure*}[th]
    \centering
    \includegraphics[width=0.95\linewidth]{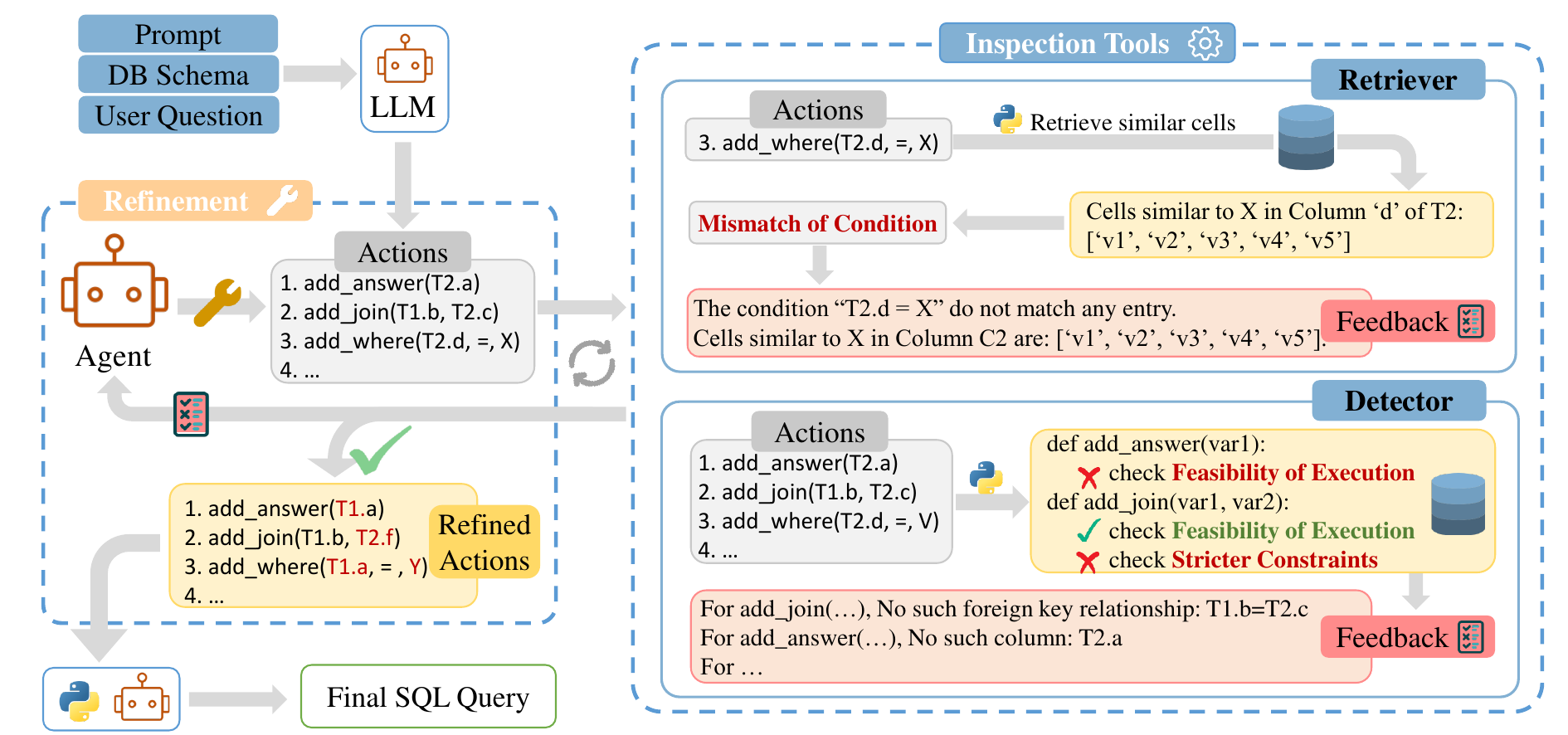}
    \caption{The overall structure of \ourmethod, consisting of an LLM-based agent for SQL refinement (left) and two tools for error inspection (right). The LLM-based agent performs actions to construct the SQL query and continuously refines these actions based on feedback from the tools. The tools target different problems in SQL queries: (i) the \textit{Database Retriever} detects the condition mismatches, and (ii) the \textit{Error Detector} diagnoses execution errors and mismatches related to stricter constraints.}
    \label{fig:method}
\end{figure*}

We introduce \ourmethod\ in Figure \ref{fig:method}, a tool-assisted agent framework designed to continuously inspect and refine SQL queries using multiple tools guided by an LLM-based agent. Following previous works \cite{jiang-etal-2023-structgpt, gu2024middleware}, we define a set of Python functions as the action space for the LLM-based agent. These functions correspond to different SQL clauses. Therefore, in our method, the output $Y$ is a sequence of actions that represent the SQL query, rather than the SQL query itself. 
With the Python interpreter executing the sequence of actions, each tool $T$ in the toolset is invoked to inspect different errors in the function calls $Y$ based on the question $Q$ and the database $D$. If an error is detected, each tool provides specific feedback $\xi$ to the LLM-based agent, helping the agent to refine specific SQL clauses rather than blindly modifying the SQL query. The inspection process can be formulated as
\begin{equation}
    \xi = T(D,Q,Y).
\end{equation}

The inspection and refinement process is iterative. After the LLM-based agent generates a sequence of actions, all the tools are invoked to inspect potential problems. If all the tools approve the action sequence, it will be used to assemble the final SQL query. In contrast, if any tool detects an issue, the agent will generate a new action sequence $Y_{i+1}$ based on the original sequence $Y_i$ and the feedback $\xi$ from the tool. 
This process may be repeated several times until all the tools approve the sequence or the maximum number of attempts is reached. The refinement process can be formulated as
\begin{equation}
    Y_{i+1} = f_{\rm agent}(I,E,S,Q,Y_i,\xi).
\end{equation}

In the following subsections, we will introduce the details of our method as follows:
\begin{itemize}
    \item The Python function calls we design for the LLM-based agent to make a sequence of actions.  
    \item The two tools integrated into our method: a database retriever and an error detector. 
    The former checks the validity of SQL conditional clauses and assists the agent by exploring database content, while the latter detects errors in queries based on SQL execution syntax, database schema, and stricter constraints defined by SQL features or domain experts.
    \item The process to obtain a final SQL query, where the Python interpreter executes an action sequence, and an LLM is employed to supplement the missing information in the SQL query.
\end{itemize}

\subsection{Function Calls Design}

\begin{figure}[t]
    \centering
    \includegraphics[width=\linewidth]{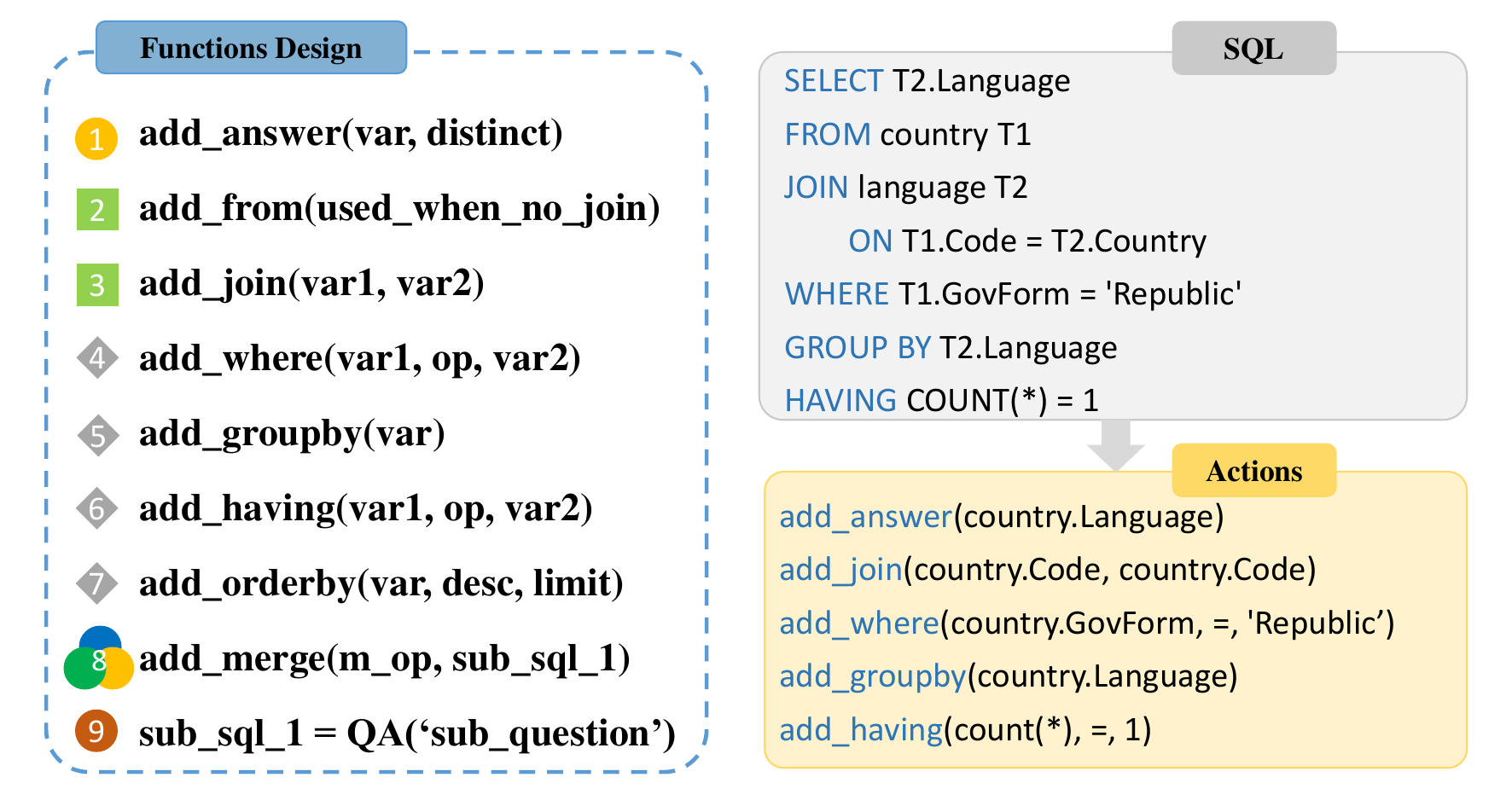}
    \caption{Python functions (left) that we designed based on the SELECT statement, along with an example (right) that includes an SQL query and the corresponding actions.}
    \label{fig:function}
\end{figure}
Figure \ref{fig:function} illustrates our function design. We define eight Python functions based on the primary clauses of the SQL, which can be used to assemble the SQL query after refinement. Each SQL clause has a corresponding function action, such as the ``WHERE" clause, which is represented by the ``add\_where" function. To reduce the action space for the LLM-based agent, we merge SQL query concatenation operators (e.g., ``UNION", ``INTERSECT", ``EXCEPT") into a single ``add\_merge" function. Additionally, conjunction operators like ``AND" and ``OR" within ``WHERE" or ``HAVING" clauses are hidden in the function design and resolved by LLM at the end, simplifying the action-making process. Since each SQL clause has distinct structural characteristics, the content of each clause is passed as parameters to the corresponding functions. For example, the ``WHERE" clause ``A = B" is represented as ``add\_where(A, =, B)". This parameterized approach allows the tools in our framework to more effectively diagnose errors in SQL clauses and reduces the complexity of string parsing. 

Besides these functions, we introduce the ``QA()" function to better address sub-questions, enhancing the reasoning ability of the agent. When this function is executed, the agent generates separate actions for sub-questions. These nine functions form the action space for the LLM Agent.

For a new question in the reasoning stage, since the context provides the LLM-based agent with a complete database schema that does not change immediately, the agent generates the sequence of actions in one go, rather than proceeding step-by-step as in other works \cite{jiang-etal-2023-structgpt,cen2024sqlfixagent}. This approach avoids excessive and unnecessary interactions between the agent and the environment. For most database table cells not covered in the context, the database retrieval tool within our framework inspects the conditional clauses and provides feedback to the agent. 

\subsection{Inspection Tools}
In this section, we define two tools---a database retriever and an error detector, which inspect problems in SQL queries and assist the LLM-based agent in refining SQL queries. 
\paragraph{Database Retriever} 
The primary responsibility of the database retriever is to assist the LLM-based agent in verifying the correctness of SQL conditional clauses. As shown in Figure \ref{fig:method}, the retriever inspects whether the parameters in conditional actions (e.g., ``add\_where" and ``add\_having") match any entry in the database and provides the agent with references to similar cells if no match is found. By using the retriever, the agent can align the value in the SQL query with the corresponding cells in the database or decide to exclude the columns from the conditional clauses, which is crucial for the Text-to-SQL task in real-world scenarios. In real-world settings, user questions often contain irregular values that differ from standardized values in the database, requiring validation before executing the query. Additionally, the ambiguity of user questions may make it challenging, even for an advanced agent, to locate the correct column names in the conditional clauses.  

We employ SimCSE \cite{gao2021simcse} as the retrieval model because it effectively captures semantic information between the constraint value and database cells, which is particularly useful for handling cases with significant variations, such as abbreviation. Based on the cells returned by the retriever, the LLM-based agent evaluates the correctness of the conditional clauses. If the condition is deemed correct, the agent selects the ground truth cell(s) to replace the mention of value; otherwise, the agent generates a new conditional action. Each time the agent makes a new action, we remove the previously used conditional columns from the database schema in the context to prevent duplicate actions. This process is repeated several times until the correct constraint is generated, or the maximum number of attempts is reached. If the maximum attempts are reached without success, we use the initial conditional action as the final answer, as it is likely to be the most accurate. 

\paragraph{Error Detector} 
The role of the error detector is to inspect the mismatches of stricter constraints and detect execution errors in SQL through indirect access to the database. SQL generated by LLM may contain errors due to factors like unfamiliarity with domain-specific SQL or hallucinations, making error detection essential. We parse the parameters of our Python functions and design verification programs to perform extensive detection with the help of databases. Different from MAC-SQL \cite{wang2024macsql}, we do not directly execute SQL queries in the DBMS to gather feedback, as this approach has limited error detection capabilities, typically only catching execution exceptions like syntax errors and database schema errors. 
Additionally, although privilege control strategies can prevent potentially harmful SQL from damaging the important data in databases, executing unchecked SQL queries is still risky, as it may lead to slow response times in large real-world databases. 

In the inspection process, we first extract the schema information of the database, including all table names, column names and types, foreign key relationships, and more. We then design verification programs to check whether the parameters of the functions satisfy the SQL operations and the database schema. For stricter constraints, our diagnostic process focuses on detecting the following errors based on SQL features: 1) mismatch of foreign key relationships, 2) redundancy or absence of ``JOIN" operations, 3) mismatch of column types in conditional clauses, 4) absence or improper usage of ``GROUP BY" clauses. We also emphasize that our tool is extensible and can be easily adapted to detect user-defined constraints in real-world scenarios by analyzing parameters in function calls. For example, in real-world scenarios where there are specific requirements on column data processing, the tool can be expanded to detect whether ``NULL" values have been excluded or whether columns with specific formats have been appropriately processed.

\subsection{SQL Generation}
In the final stage, we generate the SQL query using the corrected action sequences. We use the Python Interpreter to execute these function calls and extract the main components of the SQL query. For the missing logical operators `AND' and `OR' in `WHERE' or `HAVING' clauses, which are not included in the action sequences, we rely on the LLM to predict them. With all the components in place, we can then assemble the complete SQL query.

%% file: 3.5-Dataset.tex
\section{\ourdataset\ Dataset}
\subsection{Dataset Construction}

User questions in real-world scenarios exhibit diversity and may differ significantly from database content, especially for cells that cannot all be seen in the context of LLMs, which is not well highlighted by existing datasets. Spider \cite{yu2018spider} is a widely used benchmark for evaluating models' generalization ability across domains. Since it was built before the advent of LLMs, the utterances used in Spider questions closely resemble their paired SQL queries, i.e. the column and the values mentioned in the question are almost the same as those in gold SQL queries. Some works address the deficiencies of Spider and derive new datasets from the Spider validation set. Spider-SYN \cite{gan2021towards} modifies the schema-related utterances with corresponding synonyms. Spider-DK \cite{gan2021exploring} incorporates domain knowledge reflecting real-world question narratives. Spider-Realistic \cite{deng2021structure} paraphrases or removes explicit mentions of column names. The Bird dataset was constructed after the emergence of LLMs and focuses on more complex database content, external knowledge reasoning between problems and databases, and SQL query efficiency. 
In these datasets, values mentioned in questions are typically highlighted with initial capitalization or within quotation marks. The differences between the mention of value in questions and the ground-truth cells in databases are minimal. In most cases, they are identical, which is unlikely in real-world scenarios. In a few cases, the LLMs can infer the correct cell statements by simply seeing the first three rows of tables. Therefore, existing approaches do not account for the differences between values mentioned in questions and cells in databases, most approaches simply use the strategy of looking at a few cell examples. 

\begin{table}[t]\footnotesize
    \centering
    \renewcommand\arraystretch{0.8}
    \resizebox{\linewidth}{!}{
        \begin{tabular}{ll}
            \toprule
            Disturbance & Example \\
            \midrule
            \multirow{2}{*}{Original question} & When did the episode \\
            & ``A Love of a Lifetime" air? \\
            \midrule
            \multirow{2}{*}{Remove column}  & When did \textcolor{red}{\st{the episode}} \\
            & ``A Love of a Lifetime" air? \\
            \midrule
            \multirow{2}{*}{Remove highlight} & When did \textcolor{red}{\st{``A Love of a Lifetime"}} \\
            & \textcolor{blue}{a love of lifetime} air? \\
            \midrule
            \multirow{2}{*}{Replace common value} & When did \textcolor{red}{\st{a love of lifetime air}} \\
            & \textcolor{blue}{double down} air? \\
            \bottomrule
        \end{tabular}
    }
    \caption{Examples of the disturbances used in constructing \ourdataset\ to make the questions more realistic.}
    \label{tab:dataset_disturbances}
\end{table}

In light of the differences between value mentions in questions and cells in databases brought by the diversity of user questions in real-world scenarios, we introduce \ourdataset, a new realistic evaluation set derived from Spider, Spider-SYN, and Spider-DK. We first extract examples from these datasets which contain string-type cells in gold SQL queries, and remove duplicates and simple examples. We then manually modify both the questions and the gold SQL queries to add disturbances that reflect the real-world scenarios. The disturbances adopted are illustrated in 
Table \ref{tab:dataset_disturbances}
. While these disturbances do not introduce new SQL structures, they are expected to increase the difficulty for models in generating correct conditional clauses.

\subsection{Condition Post-processing Module}
Since existing LLM Text-to-SQL methods do not have specialized handling for values in SQL queries, generating correct conditional clauses can be challenging. To address this, we propose a simple module called Condition Post-processing. This module extracts value mentions from the predicted SQL and replaces each value with the most similar cell (retrieved using SimCSE) in the corresponding column. The condition post-process module is applied to all the methods in our experiments to provide a fair comparison.

%% file: 4-Experiments.tex
\section{Experiments}
\subsection{Experiment Setup}
\paragraph{Dataset} In addition to our dataset, we also evaluate our framework on the popular benchmark Spider \cite{yu2018spider} and Spider-Realistic \cite{deng2021structure}.

\begin{itemize}
    \item \textbf{Spider} is a large-scale dataset across 200 databases from 138 domains, aiming to assess model generalization on the unfamiliar database schema. It contains 8,659 examples in the training set, 1,034 examples in the development set, and 2,147 examples in the test set. 
    \item \textbf{Spider-Realistic} is a variant dataset of Spider, which contains 508 examples derived from the Spider development set. It disturbs the natural language questions by removing or paraphrasing the explicit mentions of column names in these questions to make them more realistic. 
\end{itemize}
\paragraph{LLM} Following previous works, we use two publicly accessible LLMs: ChatGPT (gpt-3.5-turbo-0125) and GPT-4 (gpt-4). In all experiments, we obtain model results using the official API with the temperature set to zero to ensure stable output, and max\_tokens set to 300. All other parameters of API are kept at their default settings.

\paragraph{Evaluation Metrics}
We use execution accuracy (EX) and exact match accuracy (EM), two commonly used evaluation metrics in the Text-to-SQL task, to evaluate the performance of our framework. \textit{Execution Accuracy (EX)} evaluates whether the execution results of a predicted SQL query are identical to those of the corresponding gold query. \textit{Exact Match Accuracy (EM)} requires each component of a predicted SQL query to match exactly with the gold query, though it ignores differences in the values within the SQL queries. However, since multiple correct SQL queries can exist for a given question, the EM metric may mark some correct SQL queries as incorrect. Therefore, we use execution accuracy as our primary evaluation metric. Following previous work, we utilize the evaluation script proposed by \citet{ruiqi20}.

\paragraph{Baseline}
For Spider and Spider-Realistic, we mainly choose the few-shot SOTA methods as baselines to ensure a fair comparison. Few-shot methods use only a few static examples in the context of LLMs, which are different from the demonstration selection methods that select examples from the entire training set. For \ourdataset, we compare our method with the following few-shot baselines:
\begin{itemize}
    \item {\bf DIN-SQL}: A multi-stage method that employs a self-correction approach to refine SQL queries.
    \item {\bf MAC-SQL}: A multi-agent collaboration method that refines SQL based on feedback from the DBMS. 
    \item {\bf ACT-SQL}: A single-stage method that introduces the Chain-of-Thought paradigm for SQL generation, which achieves excellent results on the Spider-Realistic dataset compared to other methods using ChatGPT.
\end{itemize}

\subsection{Main results}

\paragraph{Results on Spider and Spider-Realistic} As shown in Table \ref{tab:main_results}, \ourmethod\ achieves the highest execution accuracy on the Spider development set and the average results of the Spider development and test set. On Spider-Realistic, we achieved a larger performance gap compared to the baselines. As shown in Table \ref{tab:realistic_results}, \ourmethod\ + GPT-4 outperforms other methods that perform well on Spider by at least 4.8\%, indicating that our method is more effective at handling column disturbances in Spider-Realistic. The consistent performance of \ourmethod\ across both Spider and Spider-Realistic demonstrates its robustness in addressing challenges across different scenarios. 

\begin{table}[ht]
    \centering
    \small
    \resizebox{\linewidth}{!}{
        \begin{tabular}{l|ccc}
            \toprule
            \bf Method & \bf Dev & \bf Test & \bf Avg \\
            \midrule
            ChatGPT \cite{ouyang2022training} & 74.4 & - & - \\ 
            GPT-4 \cite{2023gpt} & 72.3 & - & - \\ 
            C3 + ChatGPT \cite{dong2023c3} & 81.2 & 82.3  & 81.9 \\ 
            ACT-SQL + ChatGPT \cite{zhang2023act} & 80.4 & - & - \\
            ACT-SQL + GPT-4 \cite{zhang2023act} & 82.9 & - & -\\
            DIN-SQL + GPT-4 \cite{pourreza2024din} & 82.8 & 85.3  & 84.5 \\
            DAIL-SQL + GPT-4 \cite{gao2023text} & 83.1 & 86.2  & 85.2 \\
            DAIL-SQL + GPT-4 + SC \cite{gao2023text} & 83.6 & \textbf{86.6} & 85.6 \\
            MAC-SQL + GPT-4 \cite{wang2024macsql} & 86.8 & 82.8  & 84.1\\
            \midrule
            \ourmethod + GPT-4 & \textbf{86.9} & 85.6 & \textbf{86.0} \\
            \bottomrule
        \end{tabular}
    }
    \caption{Execution Accuracy of \ourmethod\ and previous works on Spider. The \textit{Avg} represents the overall performance on the combined development and test set. }
    \label{tab:main_results}
\end{table}

\begin{table}[ht]\footnotesize
    \centering
    \renewcommand\arraystretch{0.8}
    \begin{tabular}{l|cc}
        \toprule
        \bf Method & \bf EX \\
        \midrule
        C3 + ChatGPT  \cite{dong2023c3}   &  75.4 \\ 
        ACT-SQL + ChatGPT \cite{zhang2023act} & 75.8 \\
        DIN-SQL + GPT-4  \cite{pourreza2024din} &  78.1 \\
        DAIL-SQL + GPT-4 \cite{gao2023text}  &  75.6 \\
        DAIL-SQL + GPT-4 + SC \cite{gao2023text} &  75.2\\
        \midrule
        \ourmethod + ChatGPT & \textbf{76.8} \\
        \ourmethod + GPT-4 & \textbf{82.9} \\
        \bottomrule
    \end{tabular}
    \caption{Execution Accuracy of \ourmethod\ and previous works on Spider-Realistic.}
    \label{tab:realistic_results}
\end{table}

\paragraph{Results on \ourdataset}
Table \ref{tab:our_results} presents the results on \ourdataset. Under the few-shot setting, \ourmethod\ exceeds the baselines by 9.6\% and 7.1\% using ChatGPT and GPT-4 as agents, respectively. \ourdataset\ targets user questions in real-world scenarios, focusing on the mismatch of conditions. Existing methods fail to extract sufficient value information from tables, making it challenging for them to generate accurate conditional clauses. \ourmethod\ maintains high performance on \ourdataset, indicating that our tools significantly enhance the capability of the LLM-based agents to handle real-world questions.

\begin{table}[t]\footnotesize
    \centering
    \renewcommand\arraystretch{0.8}
    \begin{tabular}{l|cc}
        \toprule
        \bf Method & \bf EX \\
        \midrule
        ACT-SQL + ChatGPT \cite{zhang2023act}  & 65.4 \\
        DIN-SQL + ChatGPT \cite{pourreza2024din} & 63.5 \\
        MAC-SQL + ChatGPT \cite{wang2024macsql} & 64.7 \\
        ACT-SQL + GPT-4 \cite{zhang2023act} & 73.1 \\
        DIN-SQL + GPT-4 \cite{pourreza2024din} & 78.2 \\
        MAC-SQL + GPT-4 \cite{wang2024macsql} & 74.4 \\
        \midrule
        \ourmethod + ChatGPT & \bf 75.0 \\
        \ourmethod + GPT-4 & \bf 85.3 \\
        \bottomrule
    \end{tabular}
    \caption{Execution Accuracy of \ourmethod\ and previous works on \ourdataset, which we constructed to reflect the real-world scenarios. }
    \label{tab:our_results}
\end{table}

\subsection{Ablation Study}
\begin{figure}[th]
    \centering
    \begin{subfigure}[b]{0.49\linewidth} 
        \includegraphics[width=\linewidth]{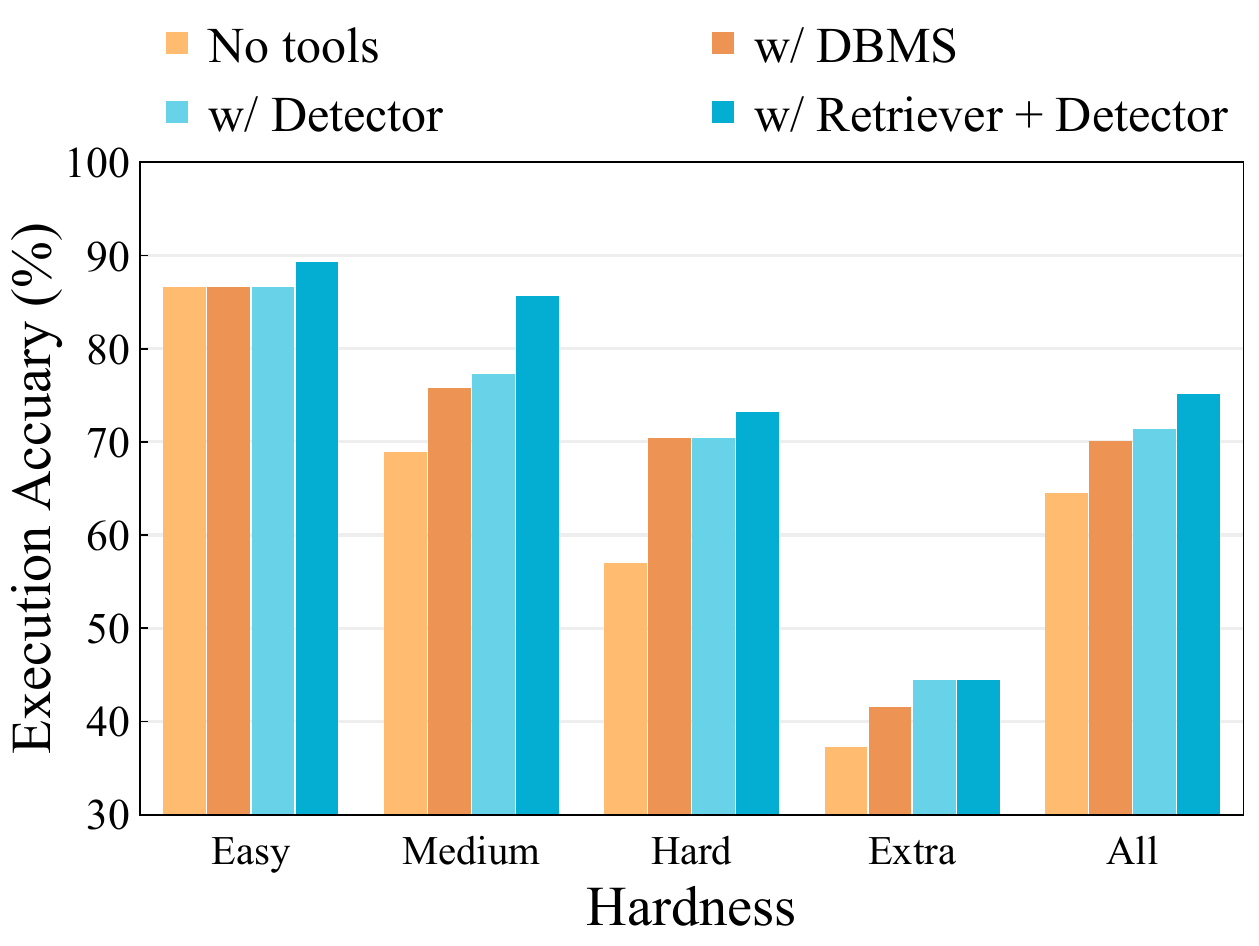}
        \captionsetup{justification=centering}
        \caption{Performance of ChatGPT}
        \label{fig:ablation_chatgpt}
    \end{subfigure}
    \begin{subfigure}[b]{0.49\linewidth}
        \includegraphics[width=\linewidth]{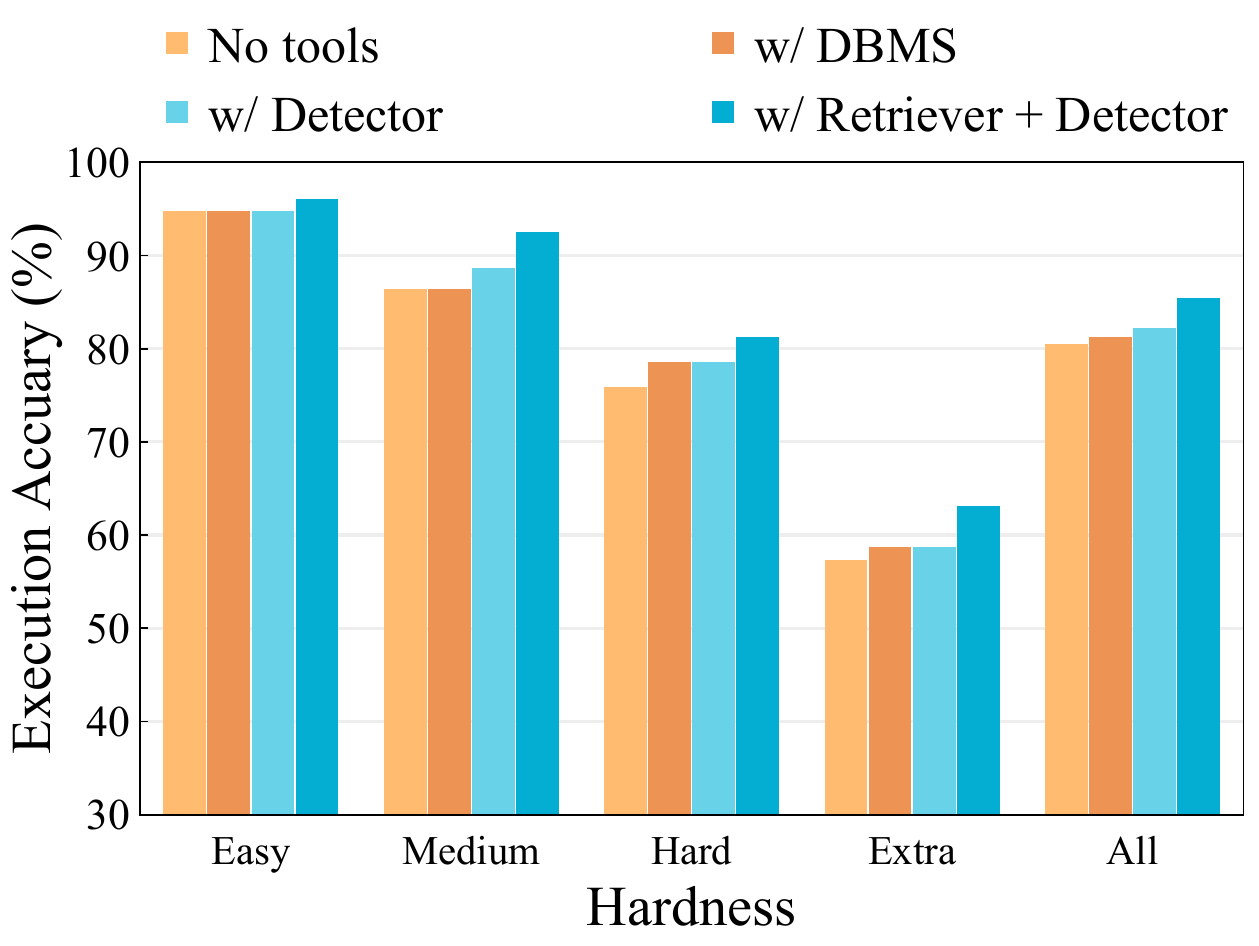}
        \captionsetup{justification=centering}
        \caption{Performance of GPT-4}
        \label{fig:ablation_gpt4}
    \end{subfigure}
    \captionsetup{aboveskip=5pt, belowskip=0pt}
    \caption{Ablation study on \ourdataset\ equipped with different tools. ``w/ Retriever + Detector" corresponds to our full method. ``w/ DBMS" denotes obtaining feedback by executing SQL queries on the database management system rather than utilizing our detector.}
    \label{fig:ablation}
\end{figure}

To evaluate the effectiveness of each verification tool in our framework, we conducted an ablation study of each tool on \ourdataset, where the questions are more reflective of real-world scenarios. As shown in Figure \ref{fig:ablation}, excluding each tool gradually decreases the performance of the LLM agent. 
When the Database Retriever is removed, the execution accuracy of ChatGPT and GPT-4 decreases by 4.1\% and 3.2\% respectively, indicating that the ambiguity of user questions makes it difficult for LLMs to generate correct SQL conditional clauses. This issue is likely to be more pronounced in real-world scenarios, where the presence of many similar cells in the database complicates the selection of the correct cell by condition post-processing methods.
Moreover, the blind application of condition post-processing methods can lead to inaccurate execution results, especially for user questions that cannot be answered by the database, highlighting the necessity of exploring cells within the database. 
Further degradation in LLM performance is observed when error detection tools are also removed, with ChatGPT showing a greater decrease. This suggests that weaker LLMs are more prone to errors without the help of the error detector. 

In addition, we replaced our error detector with database verification (corresponding to ``w/ DBMS" in Figure \ref{fig:ablation}). Compared to our error detector, the execution accuracy of ChatGPT and GPT-4 decreased by 1.6\% and 1.0\% respectively when obtaining feedback from the databases. This highlights the effectiveness of our error detector in inspecting stricter constraints for SQL queries while diagnosing the execution errors in SQL queries. 

\subsection{Discussion}

\begin{figure}[t]
    \centering
    \begin{subfigure}[b]{0.49\linewidth} 
        \includegraphics[width=\linewidth]{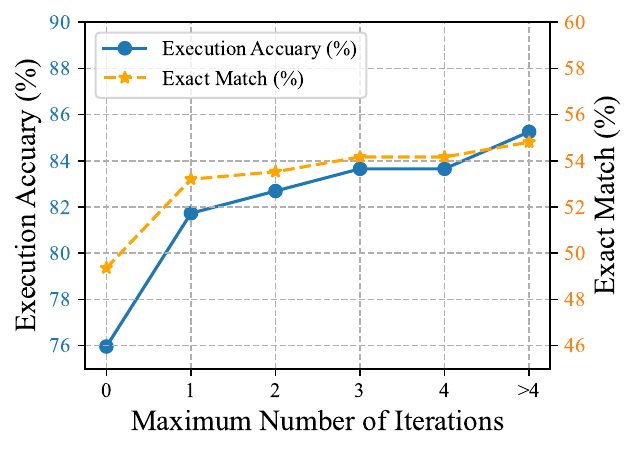}
        \captionsetup{justification=centering}
        \caption{Performance of ChatGPT}
        \label{fig:api_chatgpt}
    \end{subfigure}
    \begin{subfigure}[b]{0.49\linewidth}
        \includegraphics[width=\linewidth]{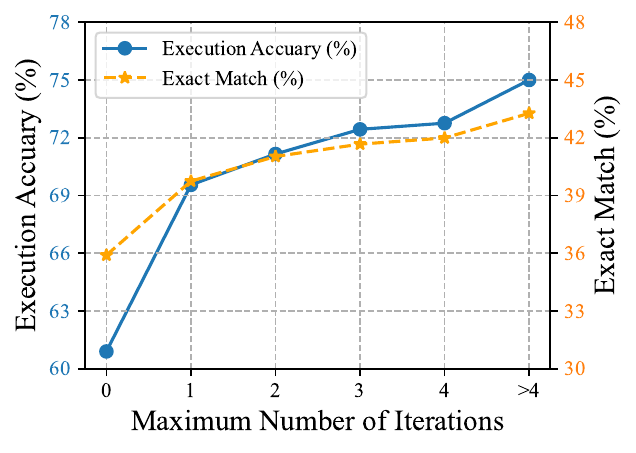}
        \captionsetup{justification=centering}
        \caption{Performance of GPT-4}
        \label{fig:api_gpt4}
    \end{subfigure}
    \captionsetup{aboveskip=5pt, belowskip=0pt}
    \caption{The impact of the maximum number of iterations on the performance of \ourmethod. }
    \label{fig:api}
\end{figure}

\paragraph{Analysis of Refinement Effectiveness}
Figure \ref{fig:api} presents the impact of the maximum number of iterations in the refinement process on \ourdataset\ using ChatGPT and GPT-4 as the agents. The results indicate that most errors in SQL queries can be solved with a single correction, mainly addressing execution errors and stricter constraint mismatches. However, there are still many errors that require multiple iterations to refine, most of which are for the refinement of conditional clauses. This suggests that LLM-based agents may need several attempts to find the correct conditions when faced with challenging user questions in real-world scenarios.

According to the results on \ourdataset, the average number of iterations required for the refinement process using \ourmethod\ + ChatGPT is 0.74, while for \ourmethod\ + GPT-4, the average is 0.44. Our method only initiates refinement when an error is detected, thereby avoiding the additional cost of introducing static processes.

\paragraph{Analysis of Condition Post-processing Module}
We also explored the impact of removing the condition post-processing module on the results. As shown in Table \ref{tab:postprocess}, the performance of all baseline methods significantly decreases after removing the post-processing module, highlighting the discrepancy between the values predicted by the LLM and the ground-truth cells in the database. 
Since our method assists LLM-based agents generate correct conditional clauses through the database detector, removing the post-processing module does not lead to a performance decline. On the contrary, introducing the module can lead to incorrect answers for the no-answer questions in real-world scenarios because the post-processing module forces the value in the conditional clauses to be replaced by the most similar cells.

\begin{table}[t]\footnotesize
    \centering
    \renewcommand\arraystretch{0.8}
    \begin{tabular}{c|c|cc}
        \toprule
        \bf Method & \bf Post-Process & \bf ChatGPT & \bf GPT-4 \\
        \midrule
        \multirow{2}{*}{ACT-SQL} & \checkmark & 65.4 & 73.1\\
        & \ding{55} & 51.9 (\textcolor{red}{13.5 $\downarrow$}) & 55.8 (\textcolor{red}{17.3 $\downarrow$})\\
        \midrule
        \multirow{2}{*}{DIN-SQL} & \checkmark & 63.5 & 78.2\\
        & \ding{55} & 33.3 (\textcolor{red}{30.2 $\downarrow$}) & 37.8 (\textcolor{red}{40.4 $\downarrow$})\\
        \midrule
        \multirow{2}{*}{MAC-SQL} & \checkmark & 64.7 & 74.4\\
        & \ding{55} & 56.1 (\textcolor{red}{8.6 $\downarrow$}) & 58.7 (\textcolor{red}{15.7 $\downarrow$})\\
        \midrule
        \ourmethod\  & \checkmark & 75.0 & 85.3 \\
        (Ours) & \ding{55} & 75.3 (\textcolor{darkgreen}{0.3 $\uparrow$}) & 85.9 (\textcolor{darkgreen}{0.6 $\uparrow$}) \\
        \bottomrule
    \end{tabular}
    \caption{Execution Accuracy of \ourmethod\ and baseline methods before and after removal of the condition post-processing module. The performance of \ourmethod\ improves after removal while other methods decline significantly.}
    \label{tab:postprocess}
\end{table}

%% file: 5-Conclusion.tex
\section{Conclusion}
In this paper, we proposed the \ourmethod\ framework designed for SQL generation in more realistic scenarios. This framework focuses on using an LLM-based agent to refine SQL queries with targeted feedback from various tools to inspect specific problems in SQL queries. 
We designed a database retriever and an error detector to address potential database mismatch problems that are common in real-world scenarios. 
The averaged experimental result on the Spider dataset and the Spider-Realistic dataset demonstrate that our method achieves the highest performance in few-shot settings. Additionally, thorough experiments on \ourdataset\ demonstrate that our method maintains high performance despite more realistic disturbances, which illustrates the effectiveness of our method in enhancing the SQL generation capabilities of LLM in real-world scenarios. 